\documentclass[final]{anthology-ch-arxiv}

\usepackage{booktabs}
\usepackage{graphicx}

\usepackage{graphicx} 
\usepackage{makecell}
\usepackage{svg}
\usepackage{hyperref}
\usepackage{tablefootnote}
\usepackage{subcaption}
\usepackage{pdfpages}

\title{Llama Nemoretriever Colembed: Top-Performing Text-Image Retrieval Model}

\pubyear{2025}
\pubvolume{1}
\pagestart{1}
\pageend{1}
\conferencename{Proceedings of Conference XXX}
\conferenceeditors{Editor1 Editor2}
\doi{00000/00000}  

\usepackage{xcolor}


\renewcommand{\pubyear}[1]{}
\renewcommand{\pubvolume}[1]{}
\renewcommand{\pagestart}[1]{}
\renewcommand{\pageend}[1]{}
\renewcommand{\conferencename}[1]{}
\renewcommand{\conferenceeditors}[1]{}
\renewcommand{\doi}[1]{}
\makeatother

\begin{document}

\maketitle

\begin{center}
\section*{Core Contributors}
Mengyao Xu, Gabriel Moreira, Ronay Ak, Radek Osmulski, Yauhen Babakhin, Zhiding Yu, Benedikt Schifferer, Even Oldridge

NVIDIA

\section*{Contributors\footnote{Sorted alphabetically}}
Adam Laiacano, Alex Richards, Andrew Tao, Ben Jarmak, Bo Liu, Charles Blackmon-Luca, Derek Whatley, Edward Kim, Fei Yu, Jeremy Dyer, Jeremy Jordan, Joey Conway, John Zedlewski, Julio Perez, Kalpesh Sutaria, Kam Mitchell, Kari Briski, Karan Sapra, Loan Luong, Maximilian Jeblick, Meghana Shrotri, Nave Algarici,  Oliver Holworthy, Padmavathy Subramanian, Randy Gelhausen, Salik Siddiqui, Sean Sodha, Shizhe Diao, Sohail Sahi, Steven Baughman, Theo Viel, Tom Balough, Tom O'Brien.

NVIDIA
\end{center}

\begin{abstract}
Motivated by the growing demand for retrieval systems that operate across modalities, we introduce llama-nemoretriever-colembed, a unified text-image retrieval model that delivers state-of-the-art performance across multiple benchmarks. We release two model variants, 1B and 3B\footnote{We released our models at \url{https://huggingface.co/nvidia/llama-nemoretriever-colembed-3b-v1} and \url{https://huggingface.co/nvidia/llama-nemoretriever-colembed-1b-v1}.\\[0.5em]}. The 3B model achieves state of the art performance, scoring NDCG@5 91.0 on ViDoRe V1 and 63.5 on ViDoRe V2, placing first on both leaderboards as of June 27, 2025.

Our approach leverages the NVIDIA Eagle2 Vision-Language model (VLM), modifies its architecture by replacing causal attention with bidirectional attention, and integrates a ColBERT-style late interaction mechanism to enable fine-grained multimodal retrieval in a shared embedding space. 
While this mechanism delivers superior retrieval accuracy, it introduces trade-offs in storage and efficiency. We provide a comprehensive analysis of these trade-offs. Additionally, we adopt a two-stage training strategy to enhance the model's retrieval capabilities.

\end{abstract}

\section{Introduction} 

Retrieval-Augmented Generation (RAG) has become a widely adopted paradigm for enhancing language models with external knowledge, enabling them to retrieve and reason on relevant content from large-scale corpora. Numerous high-performing text retrieval models including NV-Embed~\cite{nv-embed},  NV-Retriever~\cite{nv-retriever}, Qwen3-Embedding~\cite{qwen3-embed}, and e5-mistral~\cite{e5-mistral} have been proposed, and evaluated on benchmarks such as MTEB~\cite{mteb,mtebnew}. While these models and benchmarks focus primarily on text-only retrieval, they often assume clean, well-formatted textual inputs. In contrast, real-world use cases typically involve documents stored in formats like PDFs, PowerPoint slides, or Word documents, requiring preprocessing pipelines to extract textual content. This process often results in the loss of critical visual information for modalities like tables, charts, and infographics. To address this, an alternative approach proposed by ColPali~\cite{colpali} converts documents into images, enabling retrieval systems to handle both textual and visual modalities effectively.

Recent Vision-Language models (VLMs) aim to bridge the gap between text and image understanding by learning joint representations across modalities. Models such as Qwen-VL~\cite{qwenvl}, LLaMA-3.1-Nemotron-Nano-VL~\cite{nemotron}, and NVIDIA’s Eagle2 models~\cite{eagle, eagle25} have demonstrated strong performance across a range of vision-language tasks by leveraging vision encoders like CLIP~\cite{clip}, SigLIP~\cite{siglip} and C-RADIO~\cite{cradio}. CLIP and SigLIP are trained on image-text pairs using contrastive learning. C-RADIO is trained through multi-teacher distillation.
These limitations highlight the need for retrieval systems that can process documents in their native visual format, preserving both textual and visual information. This challenge has motivated the development of multimodal retrieval approaches that can understand and retrieve from documents as images.
In order to evaluate multimodal retrieval models, several benchmarks have been introduced. The most popular benchmarks on visual document retrieval are ViDoRe V1~\cite{colpali} and ViDoRe V2~\cite{vidore2}, which encompass various domains, including academic, artificial intelligence, government reports, healthcare industry, etc. 

In this report, we introduce llama-nemoretriever-colembed, a family of state-of-the-art text-image retrieval models designed for scalable and accurate multimodal retrieval. Our best-performing model, llama-nemoretriever-colembed-3b, achieves an NDCG@5 of 91.0 on ViDoRe V1 and 63.5 on ViDoRe V2, ranking No.1 on both benchmarks (as of June 27, 2025). We initialized our models from NVIDIA's Eagle2 vision-language model~\cite{eagle, eagle25}, replaced the causal attention with bidirectional attention, and fine-tuned the models through contrastive training on curated multimodal datasets. Our training datasets contain both text-only and text-image pairs, and we apply hard negative mining following the methods proposed in NV-Retriever~\cite{nv-retriever} to improve retrieval performance.
Our contributions include: 

\begin{itemize}
    \item We release two state-of-the-art text-image retrieval models: llama-nemoretriever-colembed-1B and 3B. The 3B model achieves top-1 performance on both ViDoRe V1 and ViDoRe V2 benchmarks, while the 1B model outperforms several leading 3B and 7B models.
    \item We explore two-stage training strategy, where the first stage leverages large-scale text-only data while the second stage uses text-image data. Our results demonstrate that pretraining on large-scale text-only retrieval data significantly enhances the model’s performance on downstream text-image retrieval tasks, highlighting the transferability of textual retrieval capabilities to multimodal settings.
    \item While the ColBERT-style late interaction mechanism enables fine-grained retrieval, it introduces additional overhead in terms of throughput and storage compared to simpler pooling strategies such as average or last-token pooling. We analyze and discuss the performance trade-offs of these approaches in production settings and compare an alternative method that incorporates a vision-language reranker model into the retrieval pipeline.
\end{itemize}

\section{Model}

\subsection{Model Architecture}

Our model adopts a bi-encoder retrieval framework inspired by prior dense retrieval approaches such as NV-Retriever~\cite{nv-retriever}, NV-Embed~\cite{nv-embed} and E5~\cite{e5}. In this setting, both the query and corpus items (text or image) are independently passed through a shared multimodal encoder, which projects them into a common embedding space. The relevance between a query and corpus entry is computed via a similarity function (e.g., cosine similarity or dot product), enabling fast and scalable retrieval across large corpora.

We build our retrieval model on top of the NVIDIA Eagle 2 Vision-Language Models~\cite{eagle, eagle25}.
These models adopt dynamic image tiling to support inputs of varying resolutions, and employ a carefully curated data strategy that improves multimodal learning. These design choices enable Eagle 2 models to achieve state-of-the-art results on several multimodal benchmarks, providing a solid foundation for retrieval tasks.

\begin{figure}[t!]
  \centering
  \includegraphics[width=\linewidth]{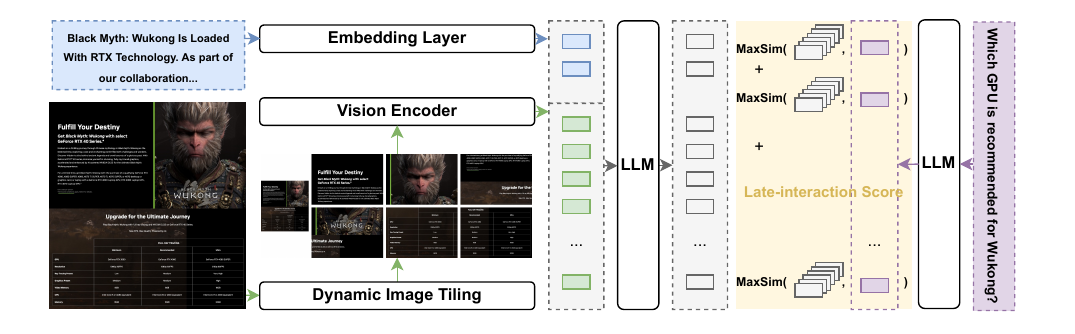}
  \caption{Multimodal Retrieval Architecture with Dynamic Image Tiling and Late-Interaction Scoring}
  \label{fig:example}
\end{figure}

We further adapt the Eagle architecture by changing the causal attention to bidirectional attention and fine-tuning it under a retrieval-specific contrastive learning objective. As part of the dynamic tiling mechanism, \texttt{max\_input\_tiles} and \texttt{min\_input\_tiles} parameters are used to control the number of tiles produced from each image. For training, we set \texttt{max\_input\_tiles = 2} to maintain memory efficiency, as increasing it to 4 showed no performance gains. During inference, we set \texttt{max\_input\_tiles = 6} to allow finer visual granularity.

To support different deployment scenarios, we develop variants of the model at multiple scales (e.g., 1B \footnote{We refer to it as the "1b" model because it leverages a 1B Llama model as the language backbone, with the complete architecture containing 2.42B parameters from both the vision encoder and language base model}, 3B) as shown in Table \ref{tab:model_arch}, allowing trade-offs between performance and retrieval accuracy.

\begin{table}[h]
  \centering 
  \begin{tabular}{ccc}
    \toprule
     Model (Huggingface ID) & Parameters (B) & Embedding Dimension\\
    \midrule
     nvidia/llama-nemoretriever-colembed-1b-v1 & 2.42 & 2048\\
     nvidia/llama-nemoretriever-colembed-3b-v1 & 4.41 & 3072 \\
    \bottomrule
  \end{tabular}
  \caption{Overview of model architecture}
  \label{tab:model_arch}
\end{table}

\subsection{Late-interaction}

\begin{figure}[ht]
  \centering
  \begin{subfigure}[t]{0.48\textwidth}
    \centering
    \includegraphics[width=\linewidth]{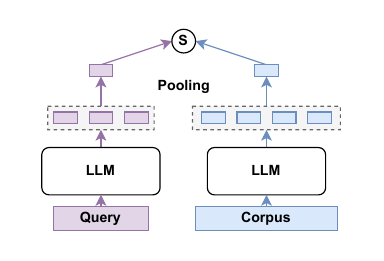}
    \caption{Bi-encoder architecture with Pooling}
    \label{fig:biencoder}
  \end{subfigure}
  \hfill
  \begin{subfigure}[t]{0.48\textwidth}
    \centering
    \includegraphics[width=\linewidth]{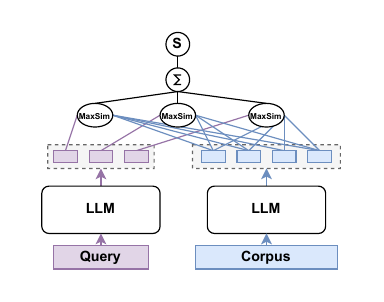}
    \caption{Late-interaction architecture}
  \label{fig:colbert}
  \end{subfigure}
  \caption{Visualization of bi-encoder and late-interaction architecture}
\end{figure}

The late interaction mechanism introduced by ColBERT~\cite{colbert} enables fine-grained interactions between query and document tokens. As shown in Figure~\ref{fig:colbert}, for a query, each token embedding interacts with all document token embeddings using a MaxSim operator, which selects the maximum similarity per query token and sums these scores to produce the final relevance score.
This requires storing all token embeddings of the document corpus (text or images).
At inference time, query token embeddings are computed and interact with the stored document embeddings through this MaxSim process. We adopt this mechanism in our models to enable fine-grained retrieval.
While this approach offers the expressiveness of token-level matching, compared to simpler pooling methods such as average or last-token pooling, as shown in Figure~\ref{fig:biencoder}, the late-interaction method introduces latency and storage overhead that may need to be assessed, as they become a concern for real-world applications. We will discuss and compare these trade-offs in Section~\ref{sec:real_world}.

\section{Training}

\subsection{Contrastive learning}

We leverage contrastive learning to maximize the embedding similarity between the query and positive passage, while minimizing the similarity between the query and negative corpus.
We adopt the InfoNCE contrastive loss~\cite{infonce} to train the model to distinguish between positive and negative pairs in a shared embedding space:

\begin{equation}
\mathcal{L}(q, d^+, D_N) = -\log \frac{\exp(\text{sim}(q, d^+)/\tau)}{\sum_{d_i \in \{d^+\} \cup D_N} \exp(\text{sim}(q, d_i)/\tau)},
\label{eq:infonce}
\end{equation}   

\noindent where $q$ is the embedding of a query, and $d^+$ are embeddings positive documents. $D_N$ denotes the set of negative passages. $ sim(\cdot) $ represents the similarity function (e.g., cosine similarity or dot product). $ \tau $ is the temperature parameter.

To improve the effectiveness of contrastive learning, we incorporate the \textit{top-k with percentage to positive threshold} strategy from NV-Retriever~\cite{nv-retriever} for hard negative mining. We set the threshold as 0.95, meaning we select the $K$ most relevant negative samples whose similarity to the query is less than 95\% of the query–positive similarity score, we set $K=2$. This encourages the model to learn from challenging negatives, while removing potential false negatives that have high similarity scores.

\subsection{Two-stage training}

\paragraph{Stage 1: Text-Only Pretraining}
In the first stage, we modify the model architecture by replacing causal attention with bidirectional attention and train it on a large-scale text-only retrieval corpora using a contrastive loss. This stage helps establish strong foundational retrieval models for textual queries and documents, allowing the model to learn semantic similarity in the embedding space.

\paragraph{Stage 2: Text-Image Fine-Tuning}
In the second stage, we fine-tune the model using text–image pairs. This stage aligns the learned text representations with visual inputs, enabling alignment across different modalities. 

\subsection{Datasets}
In the first stage, we follow the NV-Retriever methodology and use large-scale text-text pairs to establish strong language representations. Including HotpotQA~\cite{hotpotqa}, MIRACL~\cite{miracl}, Natural Questions (NQ)~\cite{nq}, Stack Exchange~\cite{stack}, SQuAD~\cite{squad} and Tiger Math/Stack~\cite{tiger}.
In the second stage, we fine-tune the model on a mixture of text–image retrieval datasets, including ColPali train set~\cite{colpali}, Wiki-SS-NQ~\cite{wikissnq, tevatron}, VDR~\cite{vdr}, VisRAG-Ret-Train-Synthetic-data~\cite{visrag}, VisRAG-Ret-Train-In-domain-data~\cite{visrag} and Docmatix~\cite{docmatix}. These datasets provide diverse and challenging multimodal examples that help the model generalize to visual retrieval tasks.
We provide a full list of training datasets in the model card on HuggingFace\footnote{https://huggingface.co/nvidia/llama-nemoretriever-colembed-3b-v1}.

\section{Results}

We evaluate our llama-nemoretriever-colembed models on two benchmarks: ViDoRe V1 and ViDoRe V2 \footnote{ViDoRe leaderboard: \url{https://huggingface.co/spaces/vidore/vidore-leaderboard}} \footnote{The results can slightly change based on the code base. ViDoRe V1 and V2 were changed to use the results based on the MTEB evaluation code. In addition, ViDoRe V2 was changed from 7 datasets to 4 datasets. Our results are based on the new ViDoRe V1 and V2 protocols, if not stated otherwise.}.
The MTEB Visual Document Retrieval (VDR) leaderboard\footnote{MTEB VDR: \url{http://mteb-leaderboard.hf.space/?benchmark_name=VisualDocumentRetrieval}} combines results from both ViDoRe benchmarks.\footnote{MTEB ranks models by Borda Count~\cite{mtebnew}, we use Avg NDCG@5 for simplicity because the Rank Borda scores are not visible in the MTEB leaderboards.} Both model variants achieve state-of-the-art performance while maintaining superior parameter efficiency.
As shown in Table~\ref{tab:vidore1new} and \ref{tab:vidore2new}, as of June 27, our llama-nemoretriever-colembed-3b-v1 model achieves top performance across both benchmarks: 91.0 on ViDoRe V1 (vs 89.9 best baseline) and 63.5 on ViDoRe V2 (vs 60.7 best baseline), resulting in a leading MTEB VDR score of 83.1 (vs 81.3 best baseline), as shown in Table~\ref{tab:doc_understanding}. The 1B variant also consistently outperforms all baselines with scores of 90.5 and 62.1 on ViDoRe V1 and V2 respectively, achieving 82.63 on MTEB VDR.
Our models demonstrate exceptional efficiency: the 1B variant (2.42B parameters) outperforms 3B and 7B baseline models, while the 3B variant (4.41B parameters) achieves state-of-the-art results using fewer parameters than 7B competitors. We provide the results for the deprecated ViDoRe V1 and V2 leaderboards in Appendix \ref{appdx:first}.


\begin{table*}[h]
  \centering
  \resizebox{\textwidth}{!}{%
    \begin{tabular}{lcccc}
      \toprule
      Model & Parameters & Embedding Dim. & Max Tokens & MTEB VDR \\
      \midrule
      
      nomic-ai/colnomic-embed-multimodal-7b
        & 7B & 128 & 128000 & 81.30 \\
      vidore/colqwen2.5-v0.2
        & 3B & 128 & 128000 & 81.23 \\
      nomic-ai/colnomic-embed-multimodal-3b
        & 3B & 128 & 128000 & 80.02 \\
      vidore/colqwen2-v1.0
        & 2B & 128 & 32768 & 79.74 \\
            vidore/colpali-v1.3
        & 2B & 128 & 16384 & 76.34 \\
      vidore/colpali-v1.2
        & 2B & 128 & 16384 & 74.72 \\
      vidore/colSmol-500M
        & 500M & 128 & 8192 & 71.36 \\
      \midrule
      \textbf{Ours} & & & & \\
      nvidia/llama-nemoretriever-colembed-1b-v1
        & 2B & 2048 & 8192 & \textbf{82.63} \\
    nvidia/llama-nemoretriever-colembed-3b-v1
        & 4B & 3072 & 8192 & \textbf{83.10} \\
      \bottomrule
    \end{tabular}
  }
  \caption{Evaluation of baseline models and our models on \href{https://huggingface.co/spaces/mteb/leaderboard}{MTEB: Visual Document Retrieval}.
Results are presented using nDCG@5 metrics.}
  \label{tab:doc_understanding}
\end{table*}

\begin{table*}[h]
  \centering
  \resizebox{\textwidth}{!}{%
    \begin{tabular}{lcccccccccccc}
      \toprule
      Model & Size (M) & Avg. & ArxivQA & DocVQA & InfoVQA & Shift Project & AI & Energy & Gov. Reports & Healthcare & TabFQuad & TAT-DQA \\
      \midrule
      
      nomic-ai/colnomic-embed-multimodal-3b
        & 3000 & 89.9 & 88.2 & 61.3 & 92.8 & 90.2 & 96.3 & 97.3 & 96.6 & 98.3 & 94.5 & 83.1 \\
      nomic-ai/colnomic-embed-multimodal-7b
        & 7000 & 89.8 & 88.4 & 60.1 & 92.3 & 89.3 & 99.3 & 96.6 & 95.4 & 99.3 & 96.1 & 81.2 \\
      vidore/colqwen2.5-v0.2
        & 3000 & 89.6 & 89.1 & 63.5 & 92.6 & 88.0 & 99.6 & 95.8 & 96.6 & 98 & 90.8 & 82.1 \\
      vidore/colqwen2-v1.0
        & 2210 & 89.2 & 88 & 61.5 & 92.5 & 89.9 & 99.0 & 95.9 & 95.5 & 98.8 & 89 & 82.2 \\
      vidore/colpali-v1.3
        & 2920 & 84.7 & 83.7 & 58.7 & 85.7 & 76.5 & 96.6 & 94.6 & 95.9 & 97.4 & 86.7 & 70.7 \\
      vidore/colpali-v1.2
        & 2920 & 83.4 & 77.9 & 56.5 & 82.4 & 78.3 & 97.5 & 94.4 & 94.9 & 95.4 & 88.4 & 68.1 \\
    \midrule
      \textbf{Ours} & & & & & & & & & & & \\
      nvidia/llama-nemoretriever-colembed-1b-v1
        & 2418 & \textbf{90.5} & 87.6 & 64.5 & 93.6 & 92.3 & 100 & 96.6 & 96.7 & 99.6 & 94.3 & 79.8 \\
    nvidia/llama-nemoretriever-colembed-3b-v1
        & 4407 & \textbf{91.0} & 88.4 & 66.2 & 94.9 & 90.7 & 99.6 & 96.6 & 97.8 & 99.3 & 95.9 & 80.6 \\
      \bottomrule
    \end{tabular}
  }
  \caption{Evaluation of baseline models and our models on \href{https://huggingface.co/spaces/vidore/vidore-leaderboard}{ViDoRe V1} (as of June 27th). Results are presented
using nDCG@5 metrics}
  \label{tab:vidore1new}
\end{table*}

\begin{table*}[h]
  \centering
  \resizebox{\textwidth}{!}{%
    \begin{tabular}{lcccccc}
      \toprule
      Model & Size (M) & Avg. & \makecell{MIT\\Biomedical\\Multilingual} & \makecell{Economics\\Macro\\Multilingual} & \makecell{ESG Restaurant\\Human\\English} & \makecell{ESG Restaurant\\Synthetic\\Multilingual} \\
      \midrule
      
      nomic-ai/colnomic-embed-multimodal-7b & 7000 & 60.7 & 63.4 & 57 & 68 & 54.4 \\
      vidore/colqwen2.5-v0.2 & 3000 & 59.5 & 59.3 & 53 & 67.1 & 58.5 \\
      nomic-ai/colnomic-embed-multimodal-3b & 3000 & 55.2 & 62.8 & 53.8 & 56.3 & 48 \\
      vidore/colpali-v1.3 & 2920 & 55.1 & 53.2 & 51 & 60.4 & 55.9 \\
      vidore/colqwen2-v1.0 & 2210 & 55 & 56.3 & 50.6 & 60.4 & 52.5 \\
      vidore/colpali-v1.2 & 2920 & 52.2 & 54.8 & 45.9 & 56.8 & 51.2 \\
      \midrule
      \textbf{Ours} & & & & & & \\      
      nvidia/llama-nemoretriever-colembed-1b-v1 & 2418 & \textbf{62.1} & 62.9 & 53.2 & 76.4 & 55.9 \\
      nvidia/llama-nemoretriever-colembed-3b-v1 & 4407 & \textbf{63.5} & 64.3 & 55.9 & 75.4 & 58.6 \\
      \bottomrule
    \end{tabular}
  }
  \caption{Evaluation of baseline models and our models on \href{https://huggingface.co/spaces/vidore/vidore-leaderboard}{ViDoRe V2} (as of June 30).
Results are presented using nDCG@5 metrics.}
  \label{tab:vidore2new}
\end{table*}

\begin{table*}[h]
  \centering
  \resizebox{\textwidth}{!}{%
    \begin{tabular}{lcccc|ccccccc}
      \toprule
      \textbf{Language} & \multicolumn{4}{c}{\textbf{MIRACL-VISION (Text)}} & \multicolumn{6}{c}{\textbf{MIRACL-VISION (Image)}} \\
      \cmidrule(lr){2-5} \cmidrule(lr){6-11}
      & \makecell{multilingual\\e5-large} & \makecell{Snowflake/snowflake\\arctic-embed\\l-v2.0} & \makecell{Alibaba-NLP/gte\\multilingual\\base} & BAAI/bge-m3 & \makecell{MrLight/dse\\qwen2-2b\\mrl-v1} & \makecell{Alibaba-NLP/gme\\Qwen2-VL-2B\\Instruct} & \makecell{llamaindex/vdr\\2b-multi-v1} & \makecell{vidore/colqwen2\\v1.0} & \makecell{nvidia/llama\\nemoretriever\\colembed-1b-v1} & \makecell{nvidia/llama\\nemoretriever\\colembed-3b-v1} \\
      \midrule
      
      Arabic & 0.8557 & 0.8754 & 0.8503 & 0.8883 & 0.3893 & \textbf{0.4888} & 0.4379 & 0.4129 & 0.3596 & 0.4247 \\
      Bengali & 0.8421 & 0.8325 & 0.8211 & 0.8585 & 0.2352 & 0.3755 & 0.2473 & 0.2888 & 0.3715 & \textbf{0.4878} \\
      Chinese & 0.6900 & 0.7179 & 0.7167 & 0.7458 & 0.5962 & \textbf{0.6314} & 0.5963 & 0.4926 & 0.3869 & 0.4355 \\
      English & 0.7029 & 0.7437 & 0.7345 & 0.7348 & 0.6605 & 0.6784 & 0.6784 & 0.6417 & 0.7165 & \textbf{0.7363} \\
      Farsi & 0.6793 & 0.7001 & 0.6984 & 0.7297 & 0.2250 & 0.3085  & 0.2398 & 0.2616 & 0.2803 & \textbf{0.3109} \\
      Finnish & 0.8974 & 0.9014 & 0.8957 & 0.9071 & 0.4162 & 0.6863 & 0.5283 & 0.6604 & 0.8278 & \textbf{0.8513} \\
      French & 0.7208 & 0.8236 & 0.7771 & 0.8158 & 0.7160 & 0.6851 & 0.7194 & 0.6876 & 0.7959 & \textbf{0.7988} \\
      German & 0.7622 & 0.7774 & 0.7498 & 0.7695 & 0.6267 & 0.6345 & 0.6205 & 0.5995 & 0.6515 & \textbf{0.6831} \\
      Hindi & 0.7595 & 0.7255 & 0.6916 & 0.7581 & 0.1740 & 0.3127 & 0.2058 & 0.2209 & 0.4670 & \textbf{0.4867} \\
      Indonesian & 0.6793 & 0.6906 & 0.6757 & 0.7049 & 0.4866 & 0.5416 & 0.5254 & 0.5320 & 0.6295 & \textbf{0.6428} \\
      Japanese & 0.8378 & 0.8484 & 0.8442 & 0.8720 & 0.6232 & \textbf{0.7305} & 0.6553 & 0.6970 & 0.6730 & 0.7260 \\
      Korean & 0.7327 & 0.7545 & 0.7397 & 0.7934 & 0.4446 & \textbf{0.6202} & 0.4952 & 0.4419 & 0.4430 & 0.5158 \\
      Russian & 0.7857 & 0.8242 & 0.8023 & 0.8363 & 0.6505 & 0.7202 & 0.6995 & 0.6811 & 0.7227 & \textbf{0.7670} \\
      Spanish & 0.6596 & 0.7250 & 0.7029 & 0.7268 & 0.5927 & 0.6277 & 0.6274 & 0.6224 & 0.7036 & \textbf{0.7109} \\
      Swahili & 0.8157 & 0.8089 & 0.7987 & 0.8337 & 0.4156 & 0.5348 & 0.4509 & 0.4931 & 0.7326 & \textbf{0.7767} \\
      Telugu & 0.8948 & 0.9201 & 0.9076 & 0.9090 & 0.0274 & 0.0893 & 0.0318 & 0.0264 & 0.0853 & \textbf{0.1669} \\
      Thai & 0.8424 & 0.8485 & 0.8509 & 0.8682 & 0.2692 & 0.3563 & 0.3177 & 0.2389 & 0.3738 & \textbf{0.4035} \\
      Yoruba & 0.5655 & 0.5332 & 0.5698 & 0.5842 & 0.4178 & 0.4884 & 0.4577 & 0.5120 & 0.5250 & \textbf{0.5888} \\
      \midrule
      AVG. & 0.7624 & 0.7806 & 0.7682 & 0.7964 & 0.4426 & 0.5283 & 0.4741 & 0.4728 & 0.5414 & \textbf{0.5841} \\
      \bottomrule
    \end{tabular}
  }
  \caption{Evaluation results on MIRACL-VISION benchmark comparing text-based and image-based retrieval models across multiple languages.}
  \label{tab:miracl_vision}
\end{table*}

MIRACL-VISION~\cite{miracl-vision} is a multilingual visual document retrieval benchmark covering 18 languages, and serves as an extension of the original MIRACL dataset. We evaluate our models on this benchmark and compare them against text-only retrieval models on the MIRACL-VISION (text) subset.
As shown in Table~\ref{tab:miracl_vision}, our models consistently outperform prior visual retrieval models across the MIRACL-VISION benchmark. The 3B variant achieves the highest overall mean score (0.5841), demonstrating strong multilingual retrieval capabilities.

\section{Real-World Applications}
\label{sec:real_world}

Leaderboards and benchmarks typically evaluate performance based on accuracy metrics. Some also include proxy indicators of computational efficiency, such as model size or embedding dimensionality. Ultimately, rankings are determined by accuracy, which may not reflect the broader needs of real-world applications. No solution fits all use-cases. In this section, we discuss the trade-offs of llama-nemoretriever-colembed in the context of production deployment.

\subsection{Review Characteristics}

Deploying a production system involves balancing accuracy, latency/throughput, and cost. A typical production system can be broken down into the following components:

\begin{itemize}
  \item \textbf{Embedding:} All embeddings of documents are generated by retrieval model. This step can be performed in batches, with support for continuous updates as new documents arrive. Throughput and cost are key considerations, and the overall retrieval performance is primarily influenced by the size of the model.
  \item \textbf{Storage:} The embeddings are stored for retrieval, and storage requirements are primarily determined by the embedding dimension and precision, i.e., how many bytes are needed per document.
  \item \textbf{Serving:} Latency measures how quickly documents can be retrieved in response to a user query. Since queries are typically short (around 50–100 tokens), the size of the embedding model plays a smaller role in this stage. Incorporating a reranker in the retrieval pipeline, such as a cross-encoder, can improve accuracy, but at the cost of increasing the latency to serve another model.

\end{itemize}

Each stage of the retrieval pipeline involves trade-offs that should be carefully aligned with the specific use case. For instance, in scenarios with a small corpus but a high volume of user queries, a larger embedding model without a reranker may offer better performance. On the other hand, for a large corpus with a moderate number of queries, a smaller embedding model combined with a reranker can be more cost-efficient. In \cite{enhancingQAranker}, we explored these trade-offs by analyzing the effects of embedding model size, inference throughput, and accuracy, both individually and with a reranker integrated into the retrieval pipeline.


\subsection{Retrieval Architecture Comparison}

ColBERT introduced the late-interaction paradigm, which demonstrated significant performance improvements in retrieval tasks by preserving fine-grained token-level interactions between queries and documents. Unlike traditional pooling strategies that compress entire sequences into single vectors, late-interaction models leverage all token-level representations. However, this approach introduces a fundamental trade-off between accuracy and storage cost, as each document requires multiple token embeddings, leading to significantly increased storage requirements.

Table~\ref{tab:performance_comparison} summarizes the comprehensive trade-offs between different retrieval approaches, reporting storage requirements in gigabytes for embedding one million images. The storage footprint of late-interaction models depends on three key factors: token count (sequence length), embedding dimension, and numerical precision (e.g., float32, float16, int8). Our analysis reveals that ColEmbed models require more storage than bi-encoder alternatives. The nvidia/llama-nemoretriever-colembed-3b-v1 model with full dimensionality (3072) requires 10,311.1 GB for one million images, representing over 2,700 times more storage than comparable bi-encoder models.

Several techniques can reduce storage requirements for both paradigms. Linear projection layers can substantially reduce embedding dimensions. Following the approach used in vidore/colqwen2-v1.0 models~\cite{colpali}, we applied linear projection layers to reduce the output dimension from 3072 to 512 and using smaller resolutions via dynamic image tiling, it reduces storage by approximately 88\% with only modest accuracy degradation (ViDoRe V1: 0.9106 to 0.9064). While this significantly decreases storage requirements to 1,230.2 GB, it still remains over 300 times larger than bi-encoder approaches.
The vidore/colqwen2-v1.0 model~\cite{colpali}, with its more compact 128-dimension embeddings, requires 179.1 GB, demonstrating how dimensionality reduction can significantly impact storage. However, bi-encoder models represent each document with a single embedding vector, requiring only 2.9-3.8 GB for one million images—at least 47 times less storage than the most compact late-interaction model in our comparison. 

Additionally, binary quantization can store embeddings using only 1-bit per element, potentially reducing storage by 16x for both architectures. However, our experience with bi-encoders indicates that binary quantization performs poorly when the embedding dimensionality is too small, and these techniques require further testing with late-interaction embedding size of 128. AnswerAI's late-pooling approach~\cite{answerai-latepooling} can reduce token vectors by factors of 3-5, while MUVERA~\cite{muvera} proposes converting multi-vector embeddings into single Fixed Dimensional Encodings (FDEs) whose inner product approximates multi-vector similarity, enabling the use of standard single-vector retrieval with smaller total embedding size.

Beyond storage, production systems must balance accuracy, latency, throughput, and inference costs. Late-interaction models face additional inference constraints due to late-interaction calculations, which require specialized vector database support and introduce latency overhead. A practical strategy is to enhance bi-encoder models with rerankers to improve retrieval accuracy while maintaining storage efficiency. Our experiments with the lama-3\_2-nemoretriever-1b-vlm-embed-v1 model~\footnote{A commercial multimodal retrieval model representing user queries as text and documents as images, \url{https://build.nvidia.com/nvidia/llama-3_2-nemoretriever-1b-vlm-embed-v1}} demonstrate this approach's effectiveness: the base model achieves ViDoRe V1: 0.8313 and V2: 0.5178 with 3.8 GB storage, while adding reranking with 25 candidates improves performance to ViDoRe V1: 0.9064 and V2: 0.6214 at the cost of approximately 2,368 ms additional latency per query.

This hybrid approach achieves performance comparable to the lama-nemoretriever-colembed-3b-v1 model while maintaining the storage advantages of bi-encoder architectures. The results demonstrate a clear trade-off between the number of reranked candidates and both inference latency and retrieval accuracy. The choice between late-interaction and bi-encoder paradigms ultimately depends on specific use case requirements and system constraints.

\begin{table*}[h]
  \centering
  \resizebox{\textwidth}{!}{%
    \begin{tabular}{llcccccccc}
      \toprule
      Architecture & Model & Avg. \# of embeddings & Embedding & \# of floating & Embedding Storage & Number & Additional Inference & ViDoRe & ViDoRe \\
      & & (Sequence Length) & Dimension & points numbers & for 1M images & Candidates & Time & V1 & V2 \\
      & & & & per document & (GB) & Reranked & (ms/query) & & \\
      \midrule
      
      ColEmbed & nvidia/llama-nemoretriever-colembed-3b-v1 & 1802 & 3072 & 5535744 & 10311.1 & N/A & N/A & 0.9106 & 0.6357 \\
      ColEmbed & nvidia/llama-nemoretriever-colembed-3b-v1 & 1290 & 512 & 660480 & 1230.2 & N/A & N/A & 0.9064 & 0.6109 \\
      ColEmbed & vidore/colqwen2-v1.0 & 751 & 128 & 96128 & 179.1 & N/A & N/A & 0.8906 & 0.5290 \\
      
      \midrule
      
      Bi-Encoder & MrLight/dse-qwen2-2b-mrl-v1 & 1 & 1536 & 1536 & 2.9 & N/A & N/A & 0.8510 & 0.5590 \\
      Bi-Encoder & lama-3\_2-nemoretriever-1b-vlm-embed-v1 & 1 & 2048 & 2048 & 3.8 & N/A & N/A & 0.8313 & 0.5178 \\
      Bi-Encoder & lama-3\_2-nemoretriever-1b-vlm-embed-v1 + reranker & 1 & 2048 & 2048 & 3.8 & 10 & 960 & 0.8931 & 0.6025 \\
      Bi-Encoder & lama-3\_2-nemoretriever-1b-vlm-embed-v1 + reranker & 1 & 2048 & 2048 & 3.8 & 25 & 2368 & 0.9064 & 0.6214 \\
      Bi-Encoder & lama-3\_2-nemoretriever-1b-vlm-embed-v1 + reranker & 1 & 2048 & 2048 & 3.8 & 100 & 9392 & 0.9101 & 0.6182 \\
      
      \bottomrule
    \end{tabular}
  }
  \caption{Comparison of system-relevant characteristics of different retrieval pipelines and models on ViDoRe benchmarks. The numbers slightly differ as we used another code base to evaluate the datasets. Sequence length is calculated by the median across ViDoRe V1. Note: There is a newer ColQwen model vidore/colqwen2.5-v0.2.}
  \label{tab:performance_comparison}
\end{table*}

\section{Conclusion}
We present llama-nemoretriever-colembed, a family of scalable and high-performing text-image retrieval models that achieve state-of-the-art results on ViDoRe V1, ViDoRe V2, and MIRACL-VISION benchmarks. By modifying the Eagle2 VLM architecture with bidirectional attention and integrating a ColBERT-style late interaction mechanism, our models support fine-grained multimodal retrieval in a shared embedding space. Trained using a two-stage training pipeline combining large-scale text and image data, our models demonstrate strong generalization and multilingual retrieval capabilities. We also analyze the trade-offs introduced by token-level late interaction and highlight key considerations for real-world deployment. Our release of both 1B and 3B model variants provides a strong foundation for future research and practical applications in multimodal retrieval.

\section*{Core Contributors}
NeMo Retriever Applied Research – Embedding and Ranking Models: Mengyao Xu, Gabriel Moreira, Ronay Ak, Radek Osmulski, Yauhen Babakhin, Benedikt Schifferer

\section*{Contributors}
NeMo Retriever Applied Research – OCR Models: Bo Liu, Theo Viel, Maximilian Jeblick

\noindent
Nemotron VLM: Zhiding Yu \footnote{Core Contributor}, Padmavathy Subramanian, Karan Sapra, Andrew Tao

\noindent
NeMo Product: Nave Algarici, Sean Sodha, Ben Jarmak

\noindent
Data: Shizhe Diao, Tom Balough

\noindent
NeMo Retriever Services:
Kalpesh Sutaria, Loan Luong, Oliver Holworthy, Jeremy Jordan, Alex Richards, Fei Yu, Salik Siddiqui, Charles Blackmon-Luca, Derek Whatley, Adam Laiacano, Tom O'Brien, Randy Gelhausen, Jeremy Dyer, Edward Kim, Sohail Sahi, Julio Perez, Steven Baughman, Kam Mitchell, Meghana Shrotri

\noindent
Management:
Even Oldridge, Joey Conway, John Zedlewski, Kari Briski


\appendix

\section{Vidore Legacy Results} \label{appdx:first}

While ViDoRe has released updated versions of their benchmarks, we include results from the previous evaluation framework for completeness. Tables~\ref{tab:vidore1old} and \ref{tab:vidore2} present our models' performance on the legacy ViDoRe V1 and V2 benchmarks, calculated using the original codebase methodology.

On the legacy ViDoRe V1 benchmark (Table~\ref{tab:vidore2}), our models maintain their competitive advantage, llama-nemoretriever-colembed-3b-v1 achieves the highest average score of 91.1, followed by our 1b variant at 90.5. Both models outperform all baseline models, with the next best performers achieving 90.4.
Similarly, on the legacy ViDoRe V2 benchmark (Table~\ref{tab:vidore2}), our models demonstrate superior performance with scores of 63.4 (3b model) and 62.6 (1b model), outperforming the best baseline of 62.1. These results are consistent with the updated benchmark evaluations.

\begin{table*}[h]
  \centering
  \resizebox{\textwidth}{!}{%
    \begin{tabular}{lccccccccccc}
      \toprule
      & Avg. & ArxivQA & DocVQA & InfoVQA & Shift Project & AI & Energy & Gov. Reports & Healthcare & TabFQuad & TAT-DQA \\
      \midrule
      tsystems/colqwen2.5-3b-multilingual-v1.0
        & 90.4 & 93.4 & 64 & 93 & 88.3 & 100 & 95.9 & 96.1 & 97.7 & 95.1 & 80.7 \\
      Metric-AI/ColQwen2.5-7b-multilingual-v1.0
        & 90.4 & 91.7 & 65.1 & 93.9 & 87.7 & 99.3 & 95.4 & 95.2 & 97.8 & 96.7 & 80.9 \\
      Metric-AI/ColQwen2.5-3b-multilingual-v1.0
        & 90.3 & 92.2 & 64.4 & 93.5 & 88.4 & 98.9 & 96.5 & 96.4 & 98.4 & 93.7 & 80.7 \\
      nomic-ai/colnomic-embed-multimodal-7b
        & 90.2 & 88.7 & 61.3 & 93.4 & 91.8 & 99.3 & 96.5 & 95.1 & 99.3 & 96 & 81.1 \\
      yydxlv/colqwen2.5-7b-v0.1
        & 90.2 & 91.1 & 63.1 & 93.5 & 89.1 & 98.9 & 95.6 & 96.2 & 98.5 & 93.9 & 81.9 \\
      tsystems/colqwen2-7b-v1.0
        & 90.1 & 90.7 & 64.5 & 92 & 89.3 & 99.3 & 96.3 & 96.3 & 99.3 & 95 & 78.6 \\
      \midrule
      \textbf{Ours} & & & & & & & & & & & \\
      nvidia/llama-nemoretriever-colembed-1b-v1 & \textbf{90.5} & 87.6 & 64.2 & 93.6 & 92.3 & 100 & 96.6 & 96.7 & 99.6 & 94.3 & 79.9 \\
      nvidia/llama-nemoretriever-colembed-3b-v1 & \textbf{91.1} & 88.4 & 65.9 & 94.9 & 90.7 & 99.6 & 96.6 & 97.8 & 99.3 & 95.9 & 80.6 \\
      \bottomrule
    \end{tabular}
  }
  \caption{Evaluation of baseline models and our models on legacy ViDoRe V1  (as of June 27).
Results are presented using nDCG@5 metrics}
  \label{tab:vidore1old}
\end{table*}

\begin{table*}[h]
  \centering
  \resizebox{\textwidth}{!}{%
    \begin{tabular}{lccccccccc}
      \toprule
      Model & Avg. & \makecell{ESG\\Restaurant\\Human} & \makecell{Economics\\Macro\\Multilingual} & \makecell{MIT\\Biomedical} & \makecell{ESG\\Restaurant\\Synthetic} & \makecell{ESG Restaurant\\Synthetic\\Multilingual} & \makecell{MIT\\Biomedical\\Multilingual} & \makecell{Economics\\Macro} \\
      \midrule
      nomic-ai/colnomic-embed-multimodal-7b & 62.1 & 73.9 & 54.7 & 66.1 & 57.3 & 56.7 & 64.2 & 61.6 \\
      vidore/colqwen2.5-v0.2 & 60.6 & 68.4 & 56.5 & 63.6 & 57.4 & 57.4 & 61.1 & 59.8 \\
      nomic-ai/colnomic-embed-multimodal-3b & 60.2 & 65.8 & 55.5 & 63.5 & 56.6 & 57.2 & 62.5 & 60.2 \\
      Alibaba-NLP/gme-Qwen2-VL-7B-Instruct & 59.3 & 65.8 & 56.2 & 64.0 & 54.3 & 56.7 & 55.1 & 62.9 \\
      nomic-ai/nomic-embed-multimodal-7b & 59.0 & 65.7 & 57.7 & 64.0 & 49.2 & 51.9 & 61.2 & 63.1 \\
      tsystems/colqwen2.5-3b-multilingual-v1.0 & 58.6 & 72.1 & 51.2 & 65.3 & 51.7 & 53.3 & 61.7 & 54.8 \\
      \midrule
      \textbf{Ours} & & & & & & & & \\
      nvidia/llama-nemoretriever-colembed-1b-v1 & \textbf{62.6} & 76.9 & 56.4 & 64.7 & 57.1 & 56.8 & 62.3 & 64.1 \\
      nvidia/llama-nemoretriever-colembed-3b-v1 & \textbf{63.4} & 74.7 & 58.0 & 65.7 & 58.8 & 57.6 & 63.2 & 66.0 \\
      \bottomrule
    \end{tabular}
  }
  \caption{Evaluation of baseline models and our models on legacy ViDoRe V2 (as of June 27).
Results are presented using nDCG@5 metrics}
  \label{tab:vidore2}
\end{table*}

\bibliographystyle{unsrt}
\bibliography{bibliography}

\end{document}